\documentclass{article}

\usepackage[preprint]{neurips_2024}

\usepackage[utf8]{inputenc} 
\usepackage[T1]{fontenc}    
\usepackage{hyperref}       
\usepackage{url}            
\usepackage{booktabs}       
\usepackage{amsfonts}       
\usepackage{amsmath}        
\usepackage{nicefrac}       
\usepackage{microtype}      
\usepackage{graphicx}       
\usepackage{xcolor}         

\title{3D Arena: An Open Platform for Generative 3D Evaluation}

\author{
  Dylan Ebert \\
  Hugging Face \\
  \texttt{dylan@huggingface.co} \\
}

\begin{document}

\maketitle

\begin{abstract}
  Evaluating Generative 3D models remains challenging due to misalignment between automated metrics and human perception of quality.
  Current benchmarks rely on image-based metrics that ignore 3D structure or geometric measures that fail to capture perceptual appeal and real-world utility.
  To address this gap, we present 3D Arena, an open platform for evaluating image-to-3D generation models through large-scale human preference collection using pairwise comparisons.
  
  Since launching in June 2024, the platform has collected 123,243 votes from 8,096 users across 19 state-of-the-art models, establishing the largest human preference evaluation for Generative 3D.
  We contribute the \texttt{iso3d} dataset of 100 evaluation prompts and demonstrate quality control achieving 99.75\% user authenticity through statistical fraud detection.
  Our ELO-based ranking system provides reliable model assessment, with the platform becoming an established evaluation resource.
  
  Through analysis of this preference data, we present insights into human preference patterns.
  Our findings reveal preferences for visual presentation features, with Gaussian splat outputs achieving a 16.6 ELO advantage over meshes and textured models receiving a 144.1 ELO advantage over untextured models.
  We provide recommendations for improving evaluation methods, including multi-criteria assessment, task-oriented evaluation, and format-aware comparison.
  The platform's community engagement establishes 3D Arena as a benchmark for the field while advancing understanding of human-centered evaluation in Generative 3D.
\end{abstract}

\section{Introduction}

The field of Generative 3D has experienced rapid growth in recent years, with state-of-the-art models emerging every few months~\cite{liu2024comprehensivesurvey3dcontent}.
These systems generate 3D assets—digital representations including meshes, point clouds, and volumetric data—that serve critical roles in downstream applications ranging from video game development and film production to architectural visualization and virtual reality experiences~\cite{akenine2019real}.
While technical capabilities have advanced rapidly, standardized evaluation methods have failed to keep pace~\cite{zhao2024challengesopportunities3dcontent}.
Current benchmarks rely predominantly on automated metrics that poorly capture real-world human preferences~\cite{fardo2016formalevaluationpsnrquality,wang2004image}, creating an evaluation gap that hinders progress in the field.

\begin{figure}[htbp]
\centering
\includegraphics[width=\textwidth]{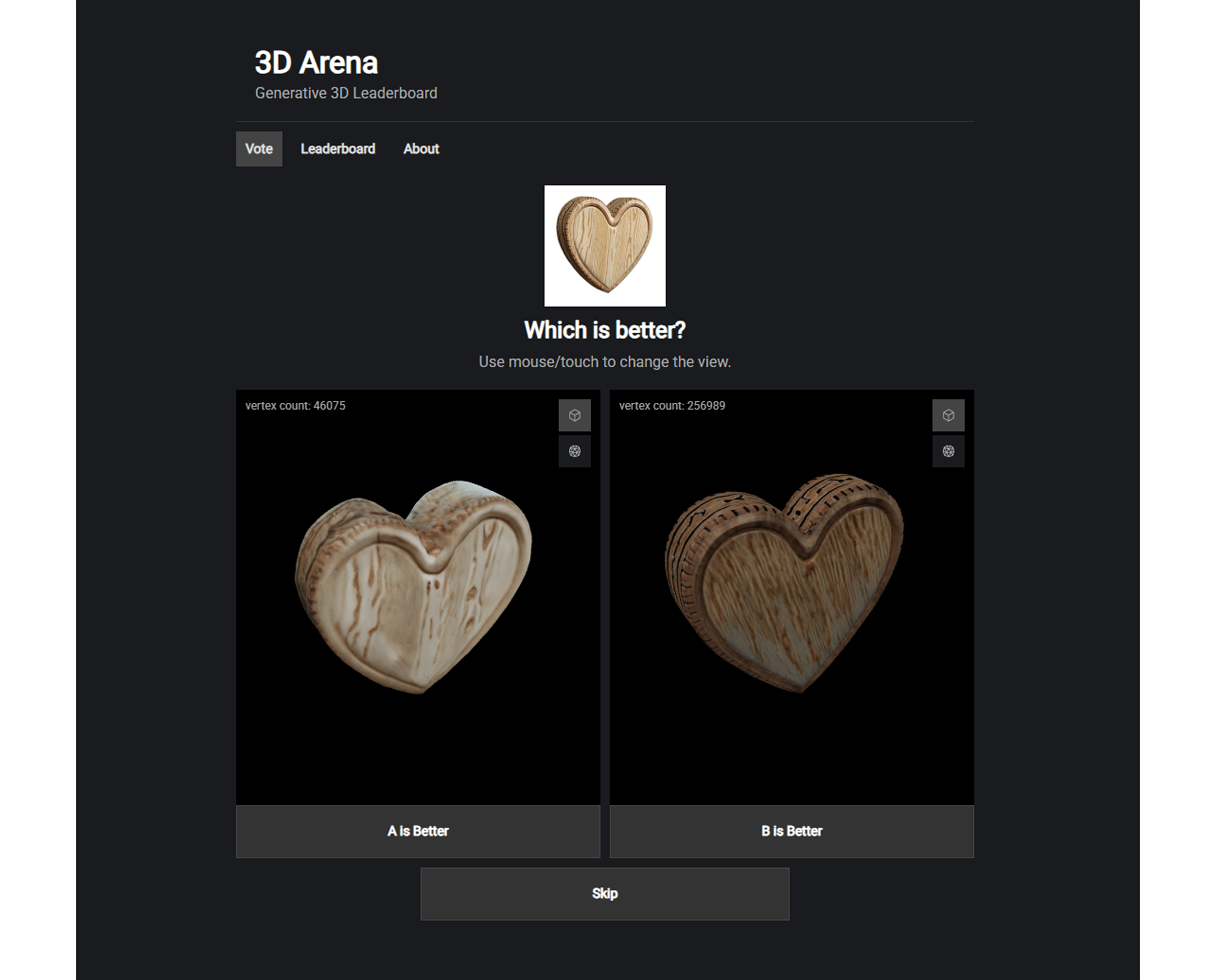}
\caption{The 3D Arena voting interface, showing side-by-side comparison of anonymized 3D outputs. Users can rotate, zoom, and examine 3D models before submitting their preference.}
\label{fig:vote}
\end{figure}

To address this challenge, we introduce 3D Arena~\footnote{\url{https://huggingface.co/spaces/dylanebert/3d-arena}}, an open platform for evaluating image-to-3D generation models through human preferences. Inspired by the success of Chatbot Arena~\cite{chiang2024chatbot},
our platform employs pairwise comparisons via an intuitive web interface where users compare anonymized 3D outputs side-by-side, as shown in Figure~\ref{fig:vote}. This approach enables direct assessment of subjective quality factors, including visual appeal, structural coherence, and perceived utility, which automated metrics fail to capture.

Since launching in June 2024, 3D Arena has established itself as the largest human preference benchmark for Generative 3D. The platform's impact extends beyond evaluation metrics, with research groups acknowledging the influence of 3D Arena rankings on their development priorities and research directions, demonstrating its practical value for guiding progress in the field.

This paper presents an analysis of 3D Arena's methods, data, and the insights we have derived from this human preference data. Our key contributions include:
\begin{enumerate}
\item \textbf{Platform validation}: Documentation of the methods used to create, operate, and maintain data integrity for crowdsourced 3D evaluation.
\item \textbf{Evaluation insights}: Analysis of human preference data revealing patterns in how humans perceive and evaluate 3D content quality.
\item \textbf{Methods framework}: Evidence-based recommendations for improving evaluation methods in Generative 3D, informed by observed preference patterns and cognitive mechanisms.
\end{enumerate}

Through analysis of this established benchmark, we provide insights for the Generative 3D community while advancing understanding of human-centered evaluation in this evolving domain.

To ensure reproducibility and enable further research, we provide open access to both the platform at \url{https://huggingface.co/spaces/dylanebert/3d-arena} and the \texttt{iso3d} evaluation dataset at \url{https://huggingface.co/datasets/dylanebert/iso3d}.

\section{Related Work}

\subsection{The Generative 3D Evaluation Challenge}

The field of Generative 3D has advanced rapidly in recent years, with state-of-the-art models emerging at an accelerating rate~\cite{zhao2024challengesopportunities3dcontent}.
Recent advances range from Shap-E~\cite{jun2023shap}
in 2023, to major leaps in quality like InstantMesh~\cite{xu2024instantmesh}
in early 2024, to systems like TRELLIS~\cite{xiang2025structured3dlatentsscalable},
Hunyuan3D-2~\cite{zhao2025hunyuan3d20scalingdiffusion},
and Hi3DGen~\cite{ye2025hi3dgenhighfidelity3dgeometry}
released in late 2024 and early 2025.
Concurrently, proprietary systems from organizations such as CSM~\cite{csm2024},
Rodin~\cite{rodin2024},
Tripo~\cite{tripo2024},
and Meshy~\cite{meshy2024}
have demonstrated high-quality Generative 3D results.

However, evaluation methods have failed to keep pace with this rapid advancement, creating a significant assessment bottleneck.
Current approaches predominantly rely on traditional image-based metrics such as PSNR~\cite{fardo2016formalevaluationpsnrquality} and SSIM~\cite{wang2004image},
which evaluate only the final rendered output while ignoring the underlying 3D structure.
While T3Bench~\cite{he2024t3benchbenchmarkingcurrentprogress}
represents an important advancement by incorporating multi-view image consistency evaluation, these image-centric approaches fail to capture the underlying 3D structure.

Attempts to address 3D structure have employed geometric metrics such as Chamfer Distance~\cite{achlioptas2018learning},
which measure shape accuracy through point-to-point distances.
However, these geometric measures fail to capture crucial aspects of 3D quality including perceptual appeal, aesthetic coherence, and mesh topology, which are factors that determine real-world utility and professional applicability.

The importance of mesh topology, in particular, has been largely overlooked by the Generative 3D research community despite its critical role in professional workflows~\cite{akenine2019real}.
Industry practitioners consistently emphasize that clean topology with proper edge flow is essential for animation, deformation, and production pipelines~\cite{parent2012computer}.
Recent work such as MeshAnything~\cite{chen2024meshanything}
addresses this gap by prioritizing mesh topology over visual fidelity.
Their evaluation approach involves user studies where participants vote on topology quality.
However, these user studies are limited by scale and cost, requiring significant effort for a small number of models and participants.

\subsection{Arena-Style Evaluation: Proven Success in Other Domains}

Arena-style evaluation has emerged as a powerful paradigm for large-scale model assessment, addressing the limitations of traditional benchmarking approaches through crowdsourced pairwise comparisons.
Chatbot Arena~\cite{chiang2024chatbot}
and GenAI Arena~\cite{jiang2024genai}
have demonstrated the effectiveness of this method, providing complementary insights to expert-designed benchmarks by capturing subjective preference aspects that automated metrics cannot assess.
Similar arena-style approaches have been applied across domains including image generation~\cite{jiang2024genai}, video generation, and text-to-image evaluation, establishing the broader applicability of community-driven preference data.

These platforms leverage well-established ranking algorithms including ELO rating systems~\cite{elo1978rating}
and the Bradley-Terry model~\cite{10.1093/biomet/39.3-4.324},
which infer relative model performance from sparse pairwise comparisons.
The core method presents users with side-by-side anonymous outputs, enabling black-box preference judgments that aggregate into statistical rankings.

The arena approach offers advantages for 3D evaluation: it captures subjective quality aspects that automated metrics cannot measure, scales to large participant populations for statistical reliability, and provides continuous assessment as new models emerge.
These characteristics make arena-style evaluation well-suited for Generative 3D, where quality encompasses multiple dimensions including visual appeal, structural coherence, and downstream utility.

Recent concurrent work has applied arena-style evaluation to Generative 3D through 3DGen-Bench~\cite{zhang20253dgen}.
While 3DGen-Bench claims to introduce "the first 3D Arena," our 3D Arena platform launched earlier in June 2024 and has since accumulated significantly more preference data with 123,000 votes compared to 3DGen-Bench's 13,800 votes.
The platforms serve complementary roles: 3D Arena functions as an open, continuously-updated leaderboard with public voting focused specifically on image-to-3D generation, while 3DGen-Bench encompasses both text-to-3D and image-to-3D evaluation through closed expert annotation with an emphasis on an automated scoring model.

\section{Methodology}

\subsection{Platform Design Principles}

3D Arena's effectiveness stems from four design principles that address limitations in existing 3D evaluation approaches:

\textbf{Anonymous Pairwise Comparison}: The platform presents users with side-by-side anonymous 3D outputs, eliminating model identity bias while enabling direct quality assessment.
Users can rotate, zoom, and examine 3D assets before submitting preferences, ensuring informed decision-making.
Model identities remain hidden until after vote submission, preventing reputation effects from influencing evaluation.

\textbf{Multi-Format Support}: The interface supports both traditional mesh formats (.obj, .glb) and emerging representations like Gaussian splats (.ply, .splat), enabling fair comparison across different Generative 3D approaches.
This design acknowledges the diversity of current Generative 3D techniques while maintaining evaluation consistency.

\textbf{Natural Preference Capture}: Rather than directing users toward specific metrics, the platform intentionally minimizes evaluation guidance, allowing natural preference patterns to emerge.
This approach reveals how users evaluate 3D content when given freedom to determine their own criteria, providing insights into real-world quality perception.

\textbf{Accessible Participation}: The platform balances scientific rigor with broad accessibility, using Hugging Face OAuth for authentication while maintaining low barriers to participation.
This design enables large-scale data collection while preserving data quality through statistical quality control.

\subsection{Evaluation Infrastructure}

\subsubsection{The \texttt{iso3d} Dataset}

To ensure fair model comparison, we present \texttt{iso3d}, a curated dataset of 100 isolated object images designed for standardized 3D evaluation.
The dataset was constructed from the Karlo-v1 prompt dataset~\cite{kakaobrain2022karlo-v1-alpha},
which contains 1,630 text prompts originally used for text-to-image model evaluation.

The generation process extended each Karlo-v1 prompt with standardizing suffixes (``isolated object render, white background'') to ensure clean, isolated objects.
Images were generated using DreamShaper-XL\footnote{\url{https://huggingface.co/Lykon/dreamshaper-xl-v2-turbo}} with White Background LoRA\footnote{\url{https://civitai.com/models/119388/white-background}}, followed by automated background removal.
From 1,630 candidates, 100 images were selected through manual review based on visual clarity and object isolation.

Physical plausibility was deliberately not controlled, including challenging inputs that test how image-to-3D models handle imperfect or physically ambiguous cases.
This design choice represents real-world usage scenarios where users input diverse images of varying quality, often with AI-generated images as intermediate steps in text-to-3D pipelines.

\subsubsection{User Interface}

\begin{figure}[htbp]
\centering
\includegraphics[width=\textwidth]{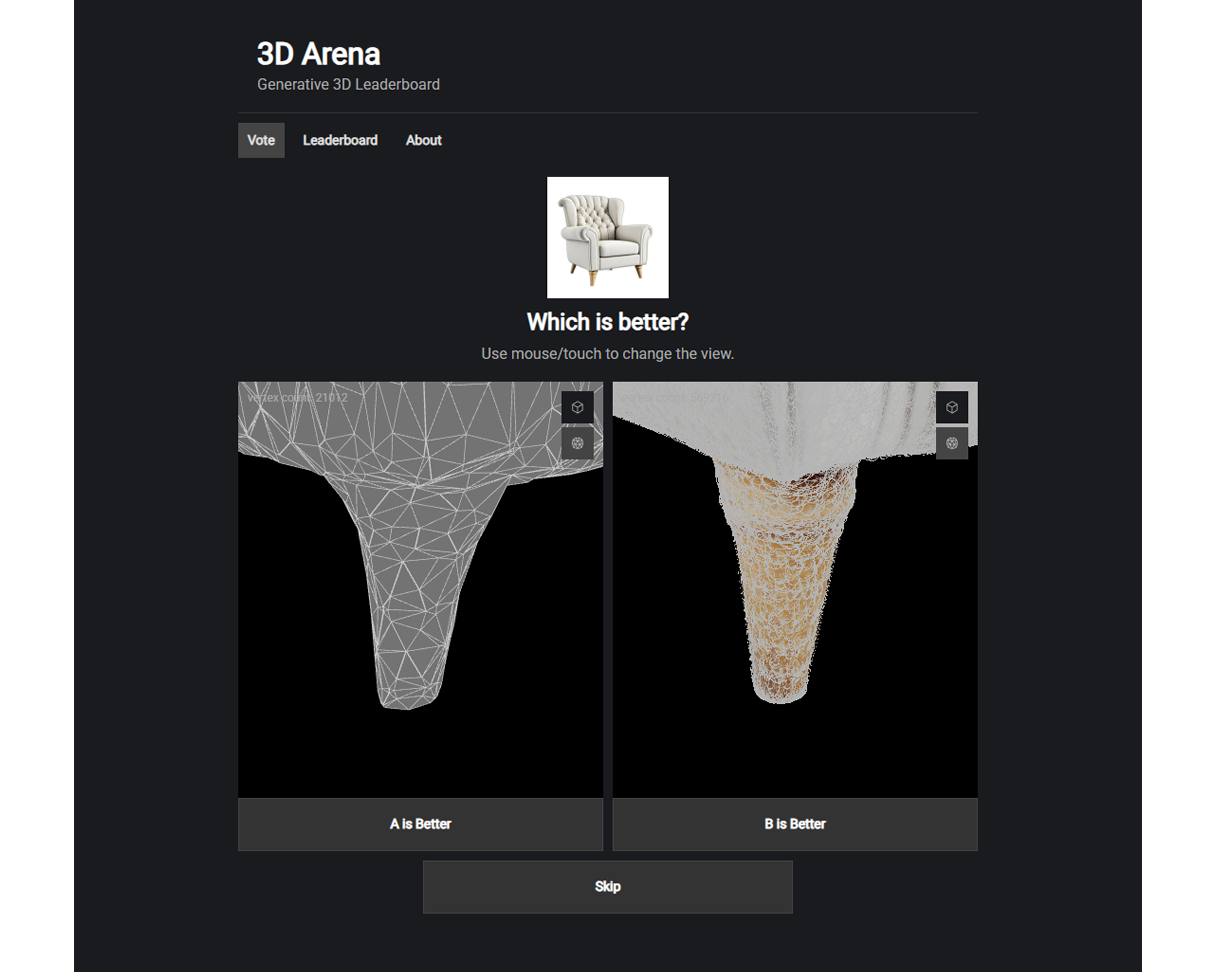}
\caption{Wireframe view of two models side-by-side showing the underlying topological differences between a mesh representation (left) and a Gaussian splat (right). This view reveals structural elements typically hidden in the standard view.}
\label{fig:topology}
\end{figure}

The platform provides side-by-side 3D viewers as shown in Figure~\ref{fig:vote}.
The standard rendered view enables evaluation of overall visual appeal and surface quality, while the wireframe view (Figure~\ref{fig:topology}) reveals underlying structural characteristics. 
Users can toggle between these two viewing modes to assess different aspects of 3D asset quality.
Additionally, polygon count is displayed for each asset.
The standard view uses default lighting and camera settings from the Gradio Model3D\footnote{\url{https://www.gradio.app/docs/gradio/model3d}} component: Babylon.js\footnote{\url{https://babylonjs.com/}} for meshes (.obj, .glb) and gsplat.js\footnote{\url{https://github.com/huggingface/gsplat.js}} for Gaussian splats (.ply, .splat).

\subsection{Quality Assurance and Statistical Methods}

3D Arena employs quality control mechanisms to ensure data integrity while preserving platform accessibility.

\textbf{User Authentication}: Users authenticate via Hugging Face OAuth to vote, associating each vote with a verified user ID.
This approach mitigates automated voting and large-scale manipulation.

\textbf{Fraud Detection}: Our statistical fraud detection system employs binomial tests comparing individual voting patterns against community consensus, flagging users whose behavior deviates significantly from expected patterns (p < 0.00001).
This approach maintains 99.75\% user authenticity, with 31 accounts flagged among 8,096 total users.

\textbf{ELO Rating System}: Model rankings utilize standard ELO rating systems, providing robust statistical estimates of relative model performance.
While platforms like Chatbot Arena employ Bradley-Terry models for enhanced statistical validity, our analysis demonstrates that ELO ratings maintain statistical robustness at our data scale, with consistent ranking patterns across evaluation periods.
All models begin with identical initial ratings (1200 points), with updates following established algorithms adapted for pairwise 3D model comparison.

\subsection{Submission and Integration Process}

3D Arena operates as an open platform designed to serve the broader research community while maintaining evaluation standards.
Model developers submit systems for evaluation through Hugging Face dataset pull requests\footnote{\url{https://huggingface.co/datasets/dylanebert/3d-arena}}, creating a transparent, community-reviewed submission process.

\textbf{Submission Standards}: Current submissions require working model implementations with consistent output format specifications.
While anonymous submissions were initially permitted and remain included in internal rankings~\ref{tab:leaderboard_all}, new submissions must provide model URLs and documentation to ensure reproducibility.
Anonymous models are excluded from the public leaderboard to maintain transparency.

\textbf{Community-Driven Evolution}: The platform evolves based on community needs and research requirements.
Submission requirements, evaluation protocols, and interface features reflect ongoing dialogue with model developers and researchers, ensuring the platform serves as a practical research tool rather than a static benchmark.

\textbf{Research Workflow Integration}: The open submission process enables rapid assessment of emerging techniques, allowing researchers to evaluate new approaches against established baselines upon development.
This integration has made 3D Arena a standard evaluation step for many research groups, establishing its role as key infrastructure for the field.

\section{Results and Findings}

\subsection{Platform Adoption and Dataset Scale}

Since launching in June 2024, 3D Arena has demonstrated widespread community adoption with \textbf{123,243 total votes} from \textbf{8,096 users} across \textbf{19 state-of-the-art models}, representing the largest scale human preference evaluation for Generative 3D.

The platform exhibits healthy participation patterns with a median of 8 votes per user and mean of 15.2 votes per user.
Vote distribution shows 61.6\% of users contributing 1-10 votes, 34.7\% contributing 11-50 votes, and 3.7\% contributing more than 50 votes.
This distribution indicates broad community engagement rather than concentration among a few "super voters," supporting the statistical validity of preference patterns.

The analyses presented in this paper are based on a snapshot of voting data from May 30, 2025, while the arena itself continues to function as a living, continuously updating benchmark with new votes and model submissions.

\subsection{Model Rankings}

Table~\ref{tab:leaderboard_all} presents the leaderboard results. 

\begin{table}[htbp]
\centering
\caption{3D Arena Leaderboard (All Models)}
\label{tab:leaderboard_all}
\begin{tabular}{@{}rlrrrr@{}}
\toprule
Rank & Model & ELO & Votes & Win Rate & Format \\
\midrule
1 & CSM/Cube~\cite{csm2024}
& 1405 & 3027 & 83.3\% & Splat \\
2 & TRELLIS-3DGS~\cite{xiang2025structured3dlatentsscalable}
& 1384 & 3648 & 80.1\% & Splat \\
3 & Strawberrry\protect\footnotemark[1] & 1382 & 4892 & 80.9\% & Mesh \\
4 & Strawb3rry\protect\footnotemark[1] & 1370 & 5121 & 79.4\% & Mesh \\
5 & TRELLIS~\cite{xiang2025structured3dlatentsscalable}
& 1306 & 4877 & 67.0\% & Mesh \\
6 & Zaohaowu3D & 1302 & 582 & 64.8\% & Mesh \\
7 & Hunyuan3D-2~\cite{zhao2025hunyuan3d20scalingdiffusion}
& 1298 & 4195 & 65.5\% & Mesh \\
8 & InstantMesh~\cite{xu2024instantmesh}
& 1278 & 10575 & 63.8\% & Mesh \\
9 & Meshy~\cite{meshy2024}
& 1243 & 7023 & 58.1\% & Mesh \\
10 & Unique3D~\cite{wu2024unique3d}
& 1230 & 8959 & 55.1\% & Mesh \\
11 & Hi3DGen~\cite{ye2025hi3dgenhighfidelity3dgeometry}
& 1207 & 1565 & 47.7\% & Mesh \\
12 & MeshFormer~\cite{liu2024meshformer}
& 1192 & 5394 & 48.6\% & Mesh \\
13 & SF3D~\cite{boss2025sf3d}
& 1190 & 6267 & 48.4\% & Mesh \\
14 & Real3D~\cite{jiang2024real3d}
& 1158 & 8541 & 42.5\% & Mesh \\
15 & SPAR3D~\cite{huang2025spar3d}
& 1144 & 3783 & 38.6\% & Mesh \\
16 & LGM~\cite{tang2024lgm}
& 1100 & 10344 & 32.7\% & Splat \\
17 & TripoSR~\cite{tochilkin2024triposr}
& 1089 & 10296 & 31.4\% & Mesh \\
18 & IM-MA~\cite{xu2024instantmesh,chen2024meshanything}\protect\footnotemark[2]
& 1016 & 7519 & 19.2\% & Mesh \\
19 & 3DTopia-XL~\cite{chen20253dtopia}
& 1000 & 5425 & 15.6\% & Mesh \\
\bottomrule
\end{tabular}
\footnotetext[1]{Anonymous submission}
\footnotetext[2]{Combination of InstantMesh generation with MeshAnything retopology}
\end{table}

The leaderboard reveals several interesting patterns in how users evaluate 3D content.

\subsubsection{Mesh vs. Splat Preferences}

3D Arena data establishes a preference for Gaussian splat outputs over mesh outputs.
Splats achieve a 16.6 ELO advantage (1215.1 vs. 1198.5), with splat formats achieving a 51.9\% win rate compared to 49.7\% for mesh formats when weighted by vote volume.
This preference pattern maintains statistical significance (p < 3.5 × 10$^{-30}$) throughout the evaluation period.

The preference occurs despite significant technical trade-offs between formats.
Gaussian splats utilize unlit rendering that appears bright and vibrant, while meshes rely on dynamic lighting models.
Splats also require substantially higher computational resources and have limited compatibility with downstream applications including animation, editing, and integration with existing 3D workflows.

A controlled comparison using TRELLIS demonstrates this format effect directly: the splat version achieves a 78 ELO advantage over its mesh counterpart despite sharing the same underlying model.

\subsubsection{Textured vs. Untextured Models}

Textured models achieve a 144.1 ELO advantage over untextured geometry (1241.1 vs. 1097.1), representing a 24.5 percentage point win rate increase (56.9\% vs. 32.4\%, p < 1.4 × 10$^{-104}$).
This aggregate pattern exhibits notable heterogeneity, with several untextured models outperforming textured counterparts.

Hi3DGen exemplifies this complexity, achieving higher ELO ratings than multiple textured models despite producing untextured meshes.
Performance variance within categories is significant (textured: 89.3 ELO standard deviation; untextured: 67.8 ELO standard deviation), indicating that texture presence alone does not determine preference outcomes.

With 74.8\% of mesh outputs incorporating textures, the data reveals that users evaluate both immediate visual appeal and structural characteristics simultaneously.
While texture provides advantages, geometric and topological factors maintain measurable influence on preference formation.
Additional controlled studies would be required to disentangle the relative contributions of these factors.

\subsubsection{Geometric Complexity Effects}

Analysis of 1,606 mesh files reveals significant variation in geometric complexity (mean: 172,571 polygons; median: 63,708).
Polygon count patterns are confounded by fundamental method differences between models.

The lowest polygon count category (<1K polygons, 1016 average ELO, 19.1\% win rate) is dominated by IM-MA, a hybrid system combining InstantMesh~\cite{xu2024instantmesh}
generation with MeshAnything~\cite{chen2024meshanything}
retopology.
IM-MA represents a topology-aware approach prioritizing mesh structure over polygon density, contrasting with topology-unaware methods across other leaderboard models.
While low polygon count is typically desirable for rendering performance and computational efficiency, IM-MA's performance reflects different optimization criteria focused on mesh topology rather than visual fidelity.

Among topology-unaware models, polygon count shows moderate positive correlation with preference (r = 0.147), with higher-density meshes (5K-20K polygons) achieving 58.8-60.9\% win rates.
The relationship exhibits diminishing returns beyond moderate complexity, indicating limited preference benefits from geometric detail within conventional generation approaches.

\subsection{Stated vs. Revealed Preferences}

Our analysis reveals a disconnect between stated preferences and observed voting behavior.
Professional 3D workflows prioritize clean topology for animation compatibility, mesh formats for standard pipelines, and technical usability for downstream applications~\cite{akenine2019real}.
Anecdotal evidence from community discourse consistently indicates similar preferences among users, with informal feedback frequently emphasizing the importance of clean mesh topology and technical usability.
However, voting patterns systematically favor visual impact through vibrant rendering and aesthetic appeal over downstream utility.

This disconnect manifests clearly in the systematic advantages of splats over meshes (16.6 ELO points) and textured over untextured models (144.1 ELO points), despite widespread industry recognition that clean mesh topology is essential for professional workflows.
This preference gap reflects established cognitive mechanisms~\cite{evans2011dual}:
the human visual system processes surface features (color, brightness) within 150-200 milliseconds~\cite{amano2006estimation},
while geometric details require additional processing time~\cite{huang2015visual}.
Features requiring deliberate evaluation are systematically underweighted compared to immediately accessible visual characteristics~\cite{orquin2013role}.

This aligns with dual-process theory from cognitive psychology~\cite{evans2011dual},
where rapid pairwise comparisons default to intuitive (System 1) evaluation, prioritizing immediate visual impressions over technical considerations requiring analytical (System 2) thinking.
The preference patterns provide evidence for the aesthetic-usability effect~\cite{kurosu1995apparent},
where aesthetically pleasing designs are perceived as more usable regardless of actual functionality.

The TRELLIS vs. TRELLIS-3DGS comparison exemplifies these mechanisms: identical underlying models achieve a 78 ELO point advantage solely through rendering differences that enhance immediate visual appeal.
Notable exceptions exist, however, with models like Hi3DGen achieving higher ratings than multiple textured alternatives despite producing untextured meshes.

\section{Discussion}

Our preference analysis reveals a gap between how 3D generation quality is perceived and what professional workflows require.
These preference patterns show that immediate visual appeal drives user preferences despite the importance of technical characteristics like clean topology for downstream applications.

This disconnect creates an optimization challenge for the field.
Model developers must balance competing objectives: creating models that satisfy user preferences while maintaining technical quality for professional utility.
The cognitive mechanisms underlying preference formation—prioritizing immediate visual impact over structural quality—favor surface features over technical characteristics that require deliberate evaluation.

3D Arena and similar platforms can address this gap in future work by implementing separate evaluation modes that isolate different quality aspects.
For topology assessment, users could be presented with only the wireframe view~\ref{fig:topology} and polygon count information, rather than rendered surfaces. These results could be used to calculate a separate \textit{topology} ELO score. This would disentangle the two quality dimensions and provide a more accurate assessment of model capabilities.

To ensure reproducibility and enable further research, we provide:

\textbf{Platform Access}: 3D Arena remains publicly accessible at \url{https://huggingface.co/spaces/dylanebert/3d-arena}, enabling continued community participation and real-time leaderboard updates.

\textbf{Dataset}: The \texttt{iso3d} evaluation dataset of 100 curated prompts is available through Hugging Face datasets at \url{https://huggingface.co/datasets/dylanebert/iso3d}, providing standardized evaluation protocols for future Generative 3D research.

\section{Conclusion}

We present 3D Arena, an open platform for evaluating Generative 3D models through large-scale human preference collection using pairwise comparisons.
Since launching in June 2024, the platform has achieved significant scale and impact, establishing the largest human preference evaluation for Generative 3D.
Through robust quality assurance and statistical fraud detection, we demonstrate that community-driven evaluation can achieve scientific reliability while preserving accessibility.

The platform has become established infrastructure for the field, with research groups integrating 3D Arena evaluation into their development workflows and acknowledging its influence on research priorities.
This practical adoption validates the platform's role as a valuable benchmark that bridges the gap between automated metrics and real-world model assessment.

Through analysis of this large-scale preference data, we contribute insights into human evaluation patterns for 3D content.
Our findings reveal systematic preferences for visual presentation features over technical characteristics, reflecting cognitive mechanisms where immediate visual features are weighted more heavily than technical aspects requiring deliberate analysis.
Importantly, these preferences capture real value for presentation-focused applications while highlighting the need for complementary evaluation approaches for technical workflows.

We provide recommendations for improving evaluation methods in Generative 3D.
Multi-criteria assessment can disentangle aesthetic appeal from technical utility, context-aware evaluation can target specific use cases, and expert vs. general user analysis can reveal how domain knowledge affects preference formation.
These approaches can provide more nuanced model assessment while preserving the valuable insights from broad community preferences.

By revealing patterns in human preference formation, our work advances understanding of quality perception in complex visual domains while providing practical guidance for model developers seeking to balance competing quality objectives.
The platform's sustained community engagement and our findings create a foundation for continued evolution of human-centered evaluation approaches in Generative 3D.

\begin{ack}
We thank the Generative 3D community for their sustained participation in 3D Arena.
Special acknowledgment goes to all model developers who have submitted their systems for evaluation and the thousands of users whose votes make this research possible.
\end{ack}

\section*{References}

\bibliographystyle{plain}
\bibliography{references}

\end{document}